# THE MULTIMODAL AND MODULAR AI CHEF: COMPLEX RECIPE GENERATION FROM IMAGERY

David Noever[1] and Samantha Elizabeth Miller Noever[2]
PeopleTec, 4901-D Corporate Drive, Huntsville, AL, USA, 35805
[1] david.noever@peopletec.com  [2] sam.igorugor@gmail.com

## ABSTRACT

*The AI community has embraced multi-sensory or multi-modal approaches to advance this generation of AI models to resemble expected intelligent understanding. Combining language and imagery represents a familiar method for specific tasks like image captioning or generation from descriptions. This paper compares these monolithic approaches to a lightweight and specialized method based on employing image models to label objects, then serially submitting this resulting object list to a large language model (LLM). This use of multiple Application Programming Interfaces (APIs) enables better than 95% mean average precision for correct object lists, which serve as input to the latest Open AI text generator (GPT-4). To demonstrate the API as a modular alternative, we solve the problem of a user taking a picture of ingredients available in a refrigerator, then generating novel recipe cards tailored to complex constraints on cost, preparation time, dietary restrictions, portion sizes, and multiple meal plans. The research concludes that monolithic multimodal models currently lack the coherent memory to maintain context and format for this task, and that until recently, the language models like GPT-2/3 struggled to format similar problems without degenerating into repetitive or non-sensical combinations of ingredients. For the first time, an AI chef or cook seems not only possible but offers some enhanced capabilities to augment human recipe libraries in pragmatic ways. The work generates a 100-page recipe book featuring the thirty top ingredients using over 2000 refrigerator images as initializing lists.*

## KEYWORDS

*Multi-modal learning, modular learning, AI model fusion, image-to-text generation, ChatGPT*

## 1. INTRODUCTION

The growing interest in combining multi-modal models [1] for AI learning algorithms has spawned several contrasting approaches. Models like Open AI's Contrastive Language-Image Pretraining (CLIP) [2], Deep Mind's Generalist Agent (GATO) [3], and LAION-5b [4] among many others fuse text and imagery into single models [5], which in their simplest usage either caption images (image-to-text) [6] or generate images from text (text-to-image) [7]. A remarkable outcome of these models centers on their capabilities for zero-shot [7] and few-shot [8] learning, such that inference stages require little or no supervised labelling of examples [9]. These approaches provide a logical pathway for combining sensory inputs and language as promising ways to achieve larger AI goals [10] than current narrow intelligence offers [11]. The example image captioning shown in Figure 1 shows a possible multimodal language facility for adding many adjectives to a valid recognition of an open refrigerator but fails to support a detailed recipe generator based on any of its current contents. The output from this image-to-text model [12] is meant to spawn another highly descriptive but synthetic image of the refrigerator's contents [13] but instead follows a more painterly approach for text output inadequate to support additional generators.

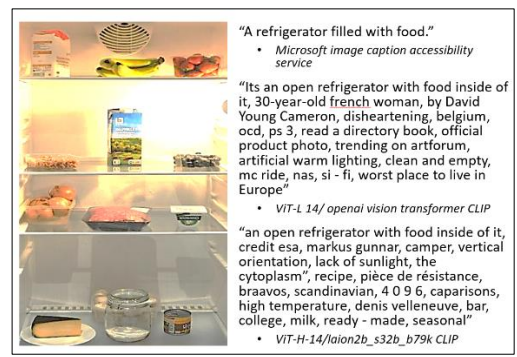

*Figure 1. Typical multimodal output without ingredient listing capabilities. Example multi-modal output from various image-to-text generators designed to caption with zero-shot labels*

One alternative approach taken here restores some reliance on tedious image labelling but benefits from letting the image and text models evolve separately [14]. The paper presents a simple application that combines image detection (e.g., food items pictured in a refrigerator) with text generation (e.g., creating a plausible recipe using as many items as possible). As an alternative to the all-inclusive, multi-modal combination of images and text, we examine a simple API pipeline approach [15]. One potential advantage to explore is the modularity of combining what each stage does best, rather than mixing training and inference. Another component to investigate stems from independently upgrading the fast-moving models in each modality, for instance, the latest text generator (like GPT-4 [16-17]) combines with convenient low-resource image models (like YoloV5 on an edge or mobile device [18-19]). The research question is empirical: demonstrate how improvements in each model might save time and computational resources in an API approach, rather than retraining a monolithic "do-anything" model of recipe suggestions. Corollaries to investigate include whether the benefits of transfer learning [20] and data augmentation [21] offers overall improved performance in this API-driven solution. The choice appears at first glance not to be mutually exclusive, particularly given the interesting, coupled problems like image captioning or visual reasoning benchmarks that necessarily require image inputs and text pre-training together.

The choice of recipe generation has an interesting history in both image and text modelling [22-51]. Text generators like the current generative pre-trained transformers (GPT-series [16-17]) combine both a creative and original layer to just standard recipe searches. In other words, recipe generators from a list of ingredients serve as interesting discovery tools for testing in collaboration with human judges [23]. In combination with image models that automate the tedious entry of ingredient lists and maintain the most current view, the API investigation provides a potentially tailored way for the image-text pipeline to take advantage of particular user needs, such as food expiration (waste reduction), dietary restrictions and preferences, and novel combinations based on available supplies. With the latest ChatGPT addition of longer prompts and memory across multiple questions [17], the user not only quickly converts a complex inventory of ingredients to automate the input [47], but also can refine the selection of choices [51] based on a guest list or seasonal menu [44].

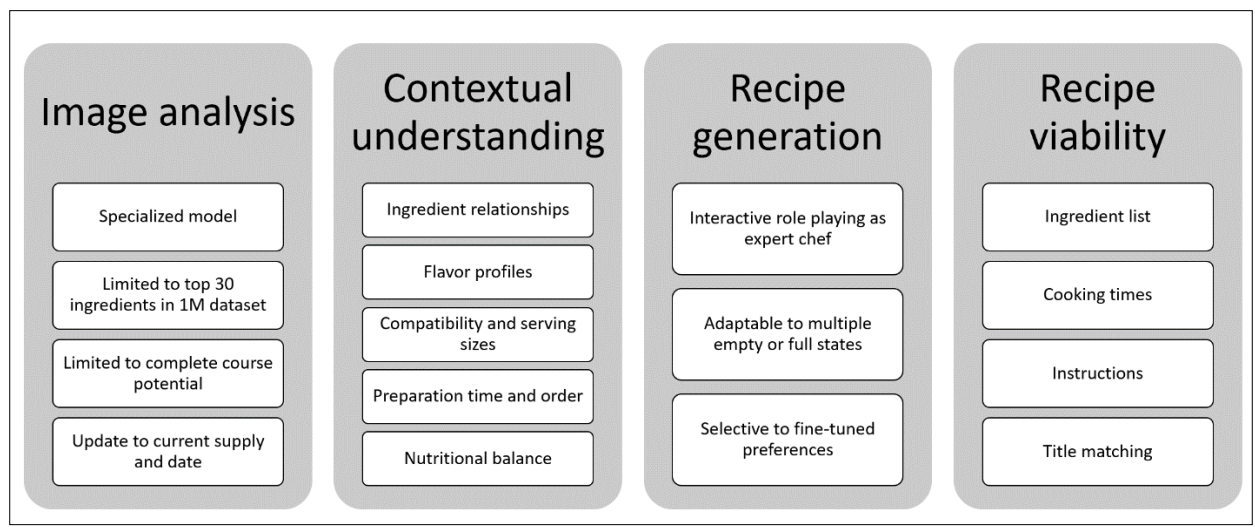

*Figure 2. Steps of modular AI cook application based on API pipeline approach*

While previous methods based on image classifiers can generate the required ingredient list, they cannot typically count the number of units (e.g., multiple servings of mushrooms) or evaluate the visual relationships needed to set a recipe's overall context (as illustrated in Figure 3). An example might be a chef's need to serve 6 people with only one chicken breast or identify longer cooking items like potatoes which may need additional preparation steps prior to serving. It is worth noting that in a combinatorial sense [52], choosing 8 ingredients as a minimum recipe input (from 30 unique classes) can generate over 38 million novel recipes. Since ingredient order does not matter and the image detector may include repetitions, the novelty formula of just ingredient combinations can be written as:

$$\# \text{ recipe combinations} = \frac{(n+r-1)!}{(n-1)!\, r!}$$

Where n is the number of items and r is the ingredients to select from the refrigerator. Of course, when one varies the cooking and preparation steps, along with the ingredient proportions, the actual number of distinct dishes would be practically endless.

Figure 2 summarize the four conceptual stages of meal planning, including the supply inventory using image detection, the context setting based on a derived ingredient list, and the tailoring of the output both to expected supply (quantities available in the refrigerator) and demand (number of required servings). The final stage of assessing the recipe viability is necessarily subjective for the current models, particularly given the size and cost of updating any large language model like Open AI's GPT-4 [17]. In the last sections of the paper, we discuss potential improvements for including human feedback or reinforcement learning in analogy to previous approaches to test kitchens [23].

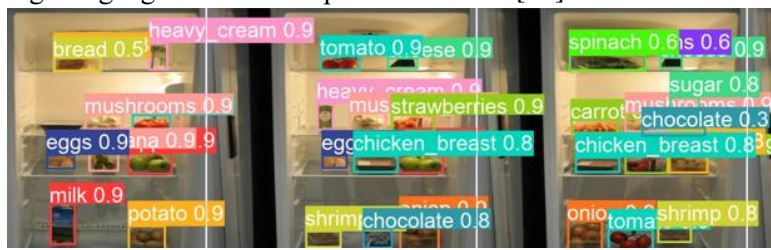

*Figure 3. Example Validation Detection and Labels for Three Images*

## 2. METHODS

To build the image detector, we trained from scratch a YoloV5 (small) model using version 4 of the open-source dataset, ai-cook-lcv4d [53]. The ai-cook dataset consists of 3,050 images of common foods (30 total) arranged in an open refrigerator. The labels localize the foods such as cheese, chicken, and apple on the refrigerator shelves. The original "objects in a fridge" project chose a subset of ingredients from the MIT CSAIL dataset of one million recipes [45], of which the ai-cook dataset further filtered to the top 50 ingredients and a final list of 30 objects to place on refrigerator shelves for photography.

As shown in Figure 4, the original 516 images (640x640 pixels) received pre-processing to auto-orient up to approximately 10-15 grocery items per image, then divided into [training: validation: testing] sets in the ratio of [2896:103:51] (or approximately [56:2:1]).

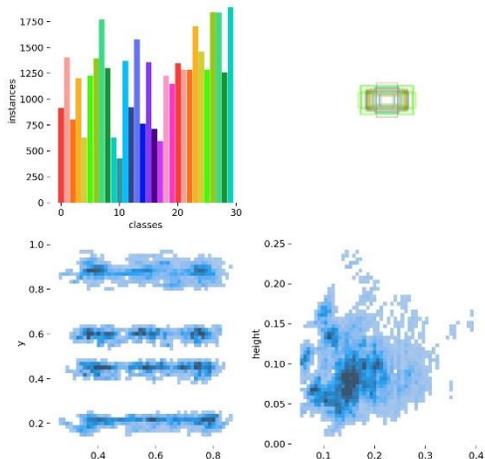

*Figure 4. Number and location distribution for 30 classes in training data*

The augmentations [21] include rotation between +/- three degrees, exposures ranging between +/-20%, blur up to 3 pixels, noise up to 5% of pixels, and cut-outs (12 boxes with 10% size each). The dataset [53] does not provide a trained example for comparison of metrics, but version 3 of the dataset yields high-quality detections on training data (mAP = 97.9%, precision = 79.6%, recall = 98.6%). It's unclear if previous authors [53] have demonstrated an equivalent validity or testing detection rate.

To train the new model, we selected YoloV5 [19] as an efficient object detector with an active user contributor base and the required frameworks in place to export trained models to the newer edge and mobile devices, particularly those having built-in cameras and access to the internet for subsequent downstream API calls. We trained a batch size of 16 over 30 epochs on a single RTX-5000 NVIDIA GPU.

To generate recipes, we collated the list of available ingredients from each image detection and submitted the available inputs to the ChatGPT3.5 (turbo) large language model (LLM) from OpenAI (March 2023) [17]. We filtered out duplicate ingredient lists where the items were rearranged on different shelves or positioned differently. Selecting from the 30 possible ingredients, each image provided from a minimum of 8 inputs and a maximum of 27 inputs to build a recipe from and translate the cooking instructions with a traditional title, ingredient list, quantities, and timed steps. The ingredient list consisted of (apple, banana, beef, blueberries, bread, butter, carrot, cheese, chicken, chicken breast, chocolate, corn, eggs, flour, goat cheese, green beans, ground beef, ham, heavy cream, lime, milk, mushrooms, onion, potato, shrimp, spinach, strawberries, sugar, sweet potato, tomato). This generation of LLMs yields complete instructions with a coherent syntax, as illustrated in Fig. 5.

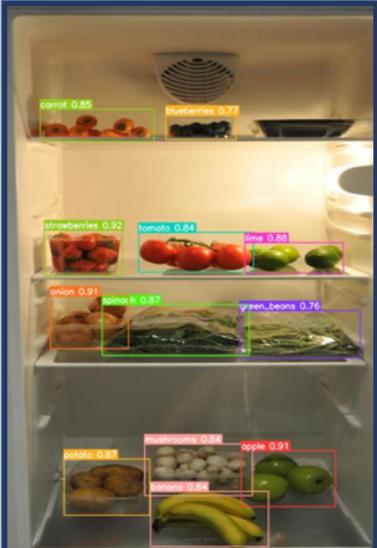

*Figure 5. Example multi-modal recipe generator based on object recognized in the fridge, then submitted to a recipe generation API*

For each list of ingredients, a common "system role" was defined to set the context of the ingredients list for GPT: *"You are an expert chef. I will give you a list of ingredients separated by a semi-colon, and you will give me a recipe using all or most of them but not supplementing outside of what might appear on this list. If an ingredient is shown as None, ignore it. Give each recipe a catchy title, an approximate time to complete and a count of people served. Also like any recipe, include the ingredient portions in the list labelled **Ingredients:** and then the **Instructions** section as a numbered list. Here's the list: <followed by a semi-colon separated list from the image detector>"*

All responses from the GPT model were zero-shot submissions, meaning whatever structural elements like ingredient proportions or cooking instructions needed to exist in the unspecialized model. For these examples, we suppressed some creative mixing of ingredients into novel combinations. The model's originality setting (model temperature=0) proves useful for a more constrained output and mimics a question-and-answer-style chatbot. The API pipeline sends a new test image to the image detector (yolov5 custom model), which generates a delimited list of ingredients ranging between 8 and 27, then submits the system context and ingredient list to the LLM for a recipe card in response. By combining the image bounding boxes and the traditional ingredient portion sizes, cooking times, and preparation stages, the overall method mimics a human chef who might spontaneously survey the kitchen state and plan an evening meal based on availability and sizes.

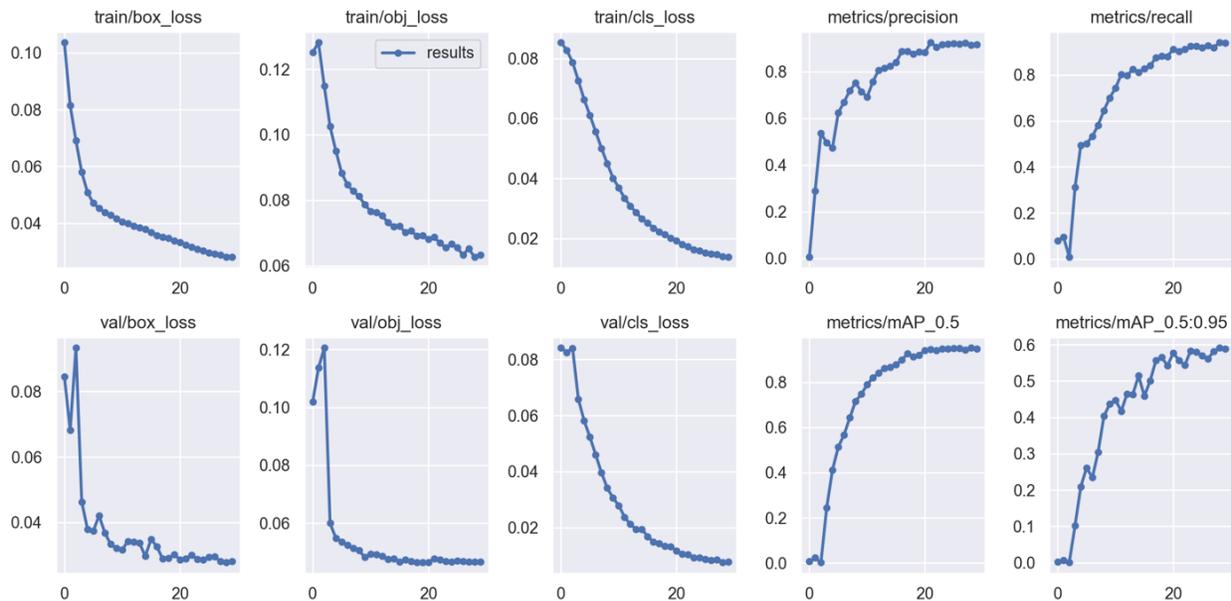

*Figure 6. Training and validation metrics for mAP, precision, and recall.*

## 3. RESULTS

Figure 6 summarizes the image detection metrics, with an overall mean average precision (mAP (0.50) =0.952 and mAP (0.5-0.95) =0.6.

In Figure 7, the multi-class confusion matrix shows that each item has a greater than 80% recognition rate with the lowest classes being "ham", "bread", and "beef" and the remainder approaching 100%. Presumably, the different shapes or packaging for these items generate some class overlap in the training dataset. The relevant output of the image detector is a delimiter-separated list of ingredients, including multiple counts of some potential ingredients to support the LLM recipe generator. A training accuracy graph for each class (Figure 7) and typical inference run for a given image (Figure 8) are shown.

The complete dataset for the approximately 100-page recipe book is hosted on Github [54].

Appendix A summarizes some additional fine-tuning examples using the interactive ChatGPT web application as a demonstration of redirecting the first recipe idea into multiple versions that might meet different human preferences for quantity, diet, cost, seasonal availability, cooking time, and multiple-meal planning.

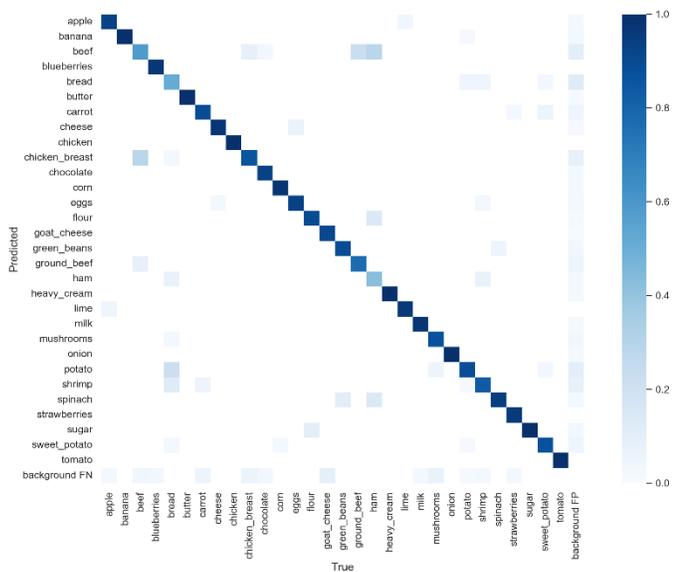

*Figure 7. Confusion matrix per class for YoloV5 and ai-cook dataset*

The long conversational memory (up to 8000 tokens or roughly 25 pages) makes possible this previously expensive interactions [17] with this generation's LLM APIs.

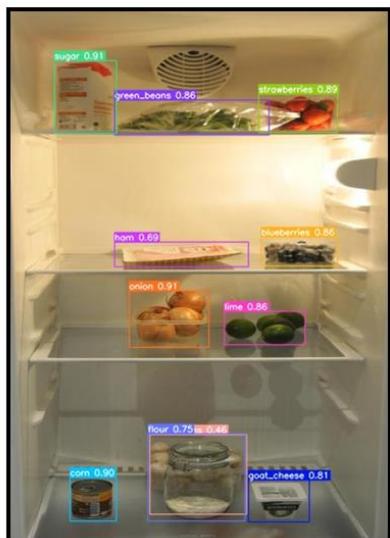

*Figure 8. The LLM and image detector combine to convert an ingredient list to a full recipe with instructions and portion sizes*

A notable outcome of using the LLM to iteratively refine the recipe for different constraints highlights the overall text flexibility and apparent conceptual comprehension of basic physical and chemical principles. For instance, to complete these recipe tasks, the text generator must demonstrate understanding basic units (Figure 8), such as the cook may not be able to split some ingredients like $1/8^{th}$ egg to bind a smaller portion of a flour tortilla. The model already demonstrates from the raw ingredients list that certain foods pair naturally with some units, like ears of corn, not ears of blueberries, or ounces of ham, not cups of hams based on the way solid ingredients are packaged vs. liquids or organic produce.

Beyond physical units, other demonstrated concepts that the LLM shows for recipe generation include understanding the seasonality of crop growth cycles (winter vs. summer vegetables). The LLM shows an ability to master the more expensive materials in a budget-trimming exercise or the long preparation times in a reduced cooking cycle. To modify cooking times, the model suggests perhaps implausible combinations given the refrigerator image, such as substituting canned corn for fresh corn or buying prepared foods like salsa rather than tomatoes.

As illustrated in Figure 9, the model modifies a baseline recipe for various dietary restrictions, including vegan, keto, and lactose-intolerant patrons. For the latter case, the recipe generated showcases a combined request to two tasks, both handling simultaneously a constraint on dietary preference (lactose intolerance) and the quantity served (2 to 3 increase in proportions).

To demonstrate basic math beyond just dividing the ingredient list by portion, the final cooking request takes the refrigerator image as its constraint but plans over time for multiple meals. In one case, a dinner for two shares the same ingredients as breakfast leftovers for one. Similarly, as a final test of the model's food reasoning, the generator must reduce a brunch menu for one person and a doggy bag for another species (poodle) with some attention to what might be harmful to the canine's digestion (onions).

Further work to spawn new conceptual or reasoning tests might carry multiple realistic scenarios encountered in modern restaurant settings. For instance, specific allergy requests now supplant many previous fast-food preparations because each meal must be customized in elaborate (often electronically entered) ingredient lists such as no lettuce, and extra sauce on the side. Beyond just seasonal constraints on supply, logistical issues at scale also serve as important elements of what now might be called operations research to menu design. For instance, a large fast-food restaurant may see a sharp increase in costs for one popular ingredient (e.g., eggs), which might incentivize the menu designer to offer a new low-calorie or low-cholesterol alternative that relies only on frozen egg whites, not fresh eggs that may be in temporary short supply or add high deliverable costs to a fixed price menu. Anecdotal evidence for AI uses within the restaurant industry [55-56] echo previous optimizations for unique items (e.g., cumin pizza) based on operations research and creative initiatives.

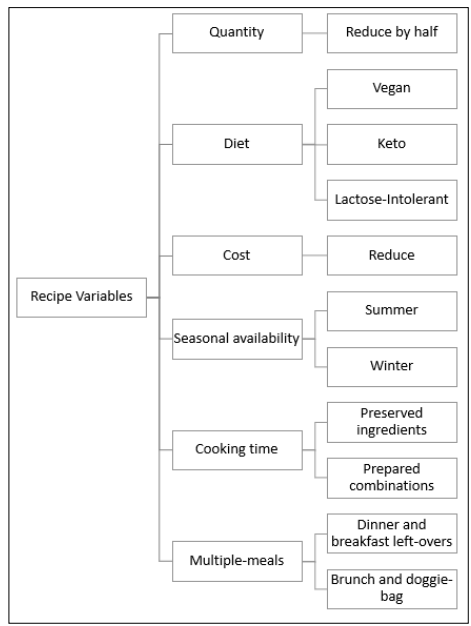

*Figure 9. Modifications to basic ingredients for specialized recipe outputs*

All these modifications seem within the capabilities of these remarkable LLMs, which as short as 2-3 years ago might struggle to duplicate the format of a recipe alone without even attempting the more conceptual understanding of what recipe generators must do as a practical matter.

## 4. DISCUSSION

The paper demonstrates a plausible API path that combines imagery and text to accomplish a highly complex and important task typically reserved for human cooks outside of the computational requirements or potential intuitive leaps and shortcuts. We trained the image model customized to a typical use case for a single (most current) image of a refrigerator filled with 30 classes of ingredients in various counts, sizes, illumination, and packaging. We passed the ingredients to a zero-shot LLM and asked for plausible recipes that might include the available ingredients, along with a relevant title, serving party size, cooking times, and ingredients list. The underlying physics and chemistry of flavor profiling or bad food combinations (bitter and sour) might provide further benchmark tests for the LLM portion of the API. A popular choice for international cuisines like Chinese, French, and Italian, modifies the same ingredients to generalize the dishes across the globe either for demonstration, contests, or seasonal events like Chinese New Year that might be celebrated within a French kitchen for a party crowd [56].

The original contribution of the research combines two model types optimized for object detection and text generation to demonstrate a highly complex human skill in a repeatable and automated fashion. While new multi-modal approaches no doubt will yield impressive fusion capabilities for describing images or painting from text descriptions, the modular API approach offers a pathway to combine state-of-the-art models that do one thing well and yet accomplish highly challenging tasks at the human expert level for both competency [44] and automated speed [49].

Further work focuses on deploying the combined image detector and text generator as an API-driven application on edge and mobile devices, along with customized user options to fine-tune or filter the resulting recipe for specific use cases. The work highlights major customization initiatives such as selecting a novel recipe for restricted diets or optimizing preparations to accommodate typical human preferences for speed, cost, seasonality, and international cuisines. One expected benefit of this approach highlights the convenience and potential to reduce environmental waste [23].

## ACKNOWLEDGMENTS

The authors benefited from the encouragement and project assistance of the PeopleTec Technical Fellows program. The authors thank the researchers at Open AI for developing large language models and allowing public access to ChatGPT.

**Appendix A: Interactive Recipe Generation (GPT-4 Model)**

**Goal:** Redirect the first recipe idea (Figure 10) into multiple versions that might meet different human preferences

**Variables:** quantity, diet, cost, seasonal availability, cooking time, and multiple-meal planning.

**System Prompt:** *You are an expert chef. I will give you a list of ingredients separated by a semi-colon, and you will give me a recipe using all or most of them but not supplementing outside of what might appear on this list. If an ingredient is shown as None, ignore it. Give each recipe a catchy title, an approximate time to complete and a count of people served. Also like any recipe, include the ingredient portions in the list labelled Ingredients: and then the Instructions section as a numbered list.*

**User Prompt (14 total ingredients):** onion, strawberries, corn, milk, sugar, cheese, banana, shrimp, goat cheese, blueberries, ground beef, flour, mushrooms

**Assistant Response:**

**Title: Surf 'n' Turf Tacos with Berry Salsa**

Approximate Time to Complete: 40 minutes
Serves: 4 people

**Ingredients:**

- 1 medium onion, finely chopped
- 1 cup strawberries, diced
- 1 cup blueberries
- 1 cup corn, fresh or frozen
- 1 cup milk
- 1 tablespoon sugar
- 1 cup shredded cheese (of your choice)
- 1 banana, sliced
- 1 pound shrimp, peeled and deveined
- 4 ounces goat cheese, crumbled
- 1 pound ground beef
- 1 cup flour
- 1 cup mushrooms, sliced
- 8 small flour tortillas

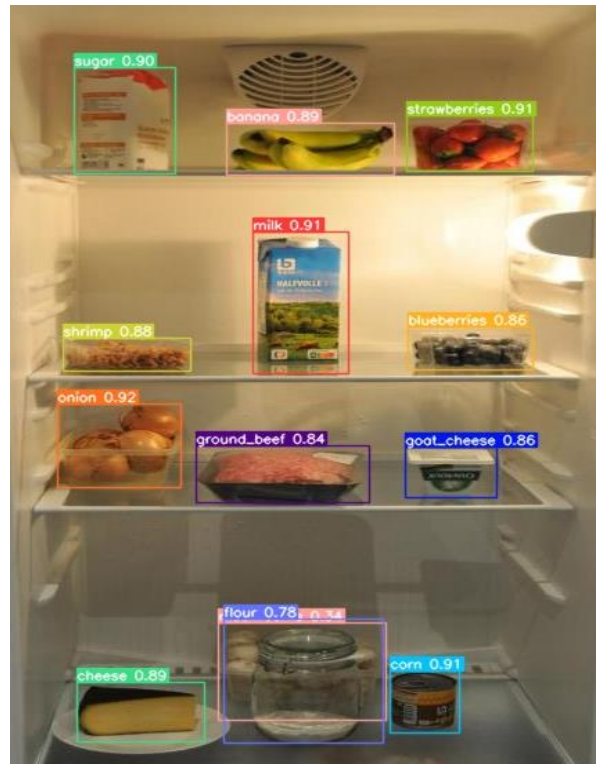

*Figure 10. Seed image for refrigerator stocks to generate the first recipe based on available identified ingredients*

**Instructions:**

1. In a large bowl, combine strawberries, blueberries, 1/2 cup of the chopped onion, and sugar. Mix well and set aside to let the flavors meld, creating the berry salsa.
2. In a large skillet, cook the ground beef over medium heat until browned. Drain the excess fat and set aside.
3. In the same skillet, cook the shrimp over medium heat until pink and cooked through. Remove from heat and set aside.

4. In a separate skillet, heat a small amount of oil over medium heat. Cook the mushrooms, remaining onion, and corn until the vegetables are tender. Set aside.
5. In a small saucepan, heat the milk over low heat. Gradually whisk in the flour to create a thick, smooth sauce. Stir in the shredded cheese until melted and combined. Remove from heat.
6. Warm the flour tortillas in a dry skillet or microwave for a few seconds until they are pliable.
7. To assemble the tacos, spread a layer of goat cheese on each tortilla. Add a portion of the ground beef, shrimp, mushroom-onion-corn mixture, and banana slices. Drizzle with the cheese sauce.

**Top the tacos with the berry salsa and serve immediately. Enjoy your Surf 'n' Turf Tacos with Berry Salsa!**

---

| |
|---|
| **User Prompt**: How would you vary the recipe if only 2 people are coming for brunch? |
| **Assistant Response**:<br>**Title:** Surf 'n' Turf Brunch Tacos with Berry Salsa (for 2)<br>Approximate Time to Complete: 40 minutes<br>Serves: 2 people<br>**Ingredients:**<br>- 1/2 medium onion, finely chopped<br>- 1/2 cup strawberries, diced<br>- 1/2 cup blueberries<br>- 1/2 cup corn, fresh or frozen<br>- 1/2 cup milk<br>- 1/2 tablespoon sugar<br>- 1/2 cup shredded cheese (of your choice)<br>- 1/2 banana, sliced<br>- 1/2 pound shrimp, peeled and deveined<br>- 2 ounces goat cheese, crumbled<br>- 1/2 pound ground beef<br>- 1/2 cup flour<br>- 1/2 cup mushrooms, sliced<br>- 4 small flour tortillas<br>**Instructions:**<br>1. In a medium bowl, combine strawberries, blueberries, 1/4 cup of the chopped onion, and sugar. Mix well and set aside to let the flavors meld, creating the berry salsa.<br>2. In a large skillet, cook the ground beef over medium heat until browned. Drain the excess fat and set aside.<br>3. In the same skillet, cook the shrimp over medium heat until pink and cooked through. Remove from heat and set aside.<br>4. In a separate skillet, heat a small amount of oil over medium heat. Cook the mushrooms, remaining onion, and corn until the vegetables are tender. Set aside.<br>5. In a small saucepan, heat the milk over low heat. Gradually whisk in the flour to create a thick, smooth sauce. Stir in the shredded cheese until melted and combined. Remove from heat.<br>6. Warm the flour tortillas in a dry skillet or microwave for a few seconds until they are pliable.<br>7. To assemble the tacos, spread a layer of goat cheese on each tortilla. Add a portion of the ground beef, shrimp, mushroom-onion-corn mixture, and banana slices. Drizzle with the cheese sauce.<br>8. Top the tacos with the berry salsa and serve immediately. Enjoy your Surf 'n' Turf Brunch Tacos with Berry Salsa for two! |
| **User Prompt**: Nice. How would you modify the recipe to minimize the meal's production costs? |
| **Assistant Response**:<br>**Title:** Budget-Friendly Surf 'n' Turf Tacos with Simple Salsa<br>Approximate Time to Complete: 35 minutes<br>Serves: 2 people |

**Ingredients:**
- 1/2 medium onion, finely chopped
- 1/2 cup corn, fresh or frozen
- 1/2 cup milk
- 1/2 tablespoon sugar
- 1/2 cup shredded cheese (of your choice, preferably a more affordable variety)
- 1/2 pound ground beef (or ground turkey or chicken for a more cost-effective alternative)
- 1/2 cup flour
- 1/2 cup mushrooms, sliced
- 4 small flour tortillas

Simple Salsa:
- 1 cup diced tomatoes
- 1/4 cup finely chopped onion
- 1/2 jalapeño, seeded and finely chopped (optional)
- Salt and pepper, to taste

**Instructions:**
1. In a medium bowl, combine diced tomatoes, 1/4 cup of the chopped onion, jalapeño (if using), salt, and pepper. Mix well and set aside to let the flavors meld, creating the simple salsa.
2. In a large skillet, cook the ground beef (or ground turkey or chicken) over medium heat until browned. Drain the excess fat and set aside.
3. In the same skillet, heat a small amount of oil over medium heat. Cook the mushrooms, remaining onion, and corn until the vegetables are tender. Set aside.
4. In a small saucepan, heat the milk over low heat. Gradually whisk in the flour to create a thick, smooth sauce. Stir in the shredded cheese until melted and combined. Remove from heat.
5. Warm the flour tortillas in a dry skillet or microwave for a few seconds until they are pliable.
6. To assemble the tacos, spread a thin layer of the cheese sauce on each tortilla. Add a portion of the ground meat, mushroom-onion-corn mixture, and drizzle with a bit more cheese sauce.
7. Top the tacos with the simple salsa and serve immediately. Enjoy your Budget-Friendly Surf 'n' Turf Tacos with Simple Salsa!

**Modifications to minimize production costs:**
- Removed shrimp, goat cheese, strawberries, blueberries, and banana to reduce the cost of ingredients.
- Replaced the original berry salsa with a simpler, more affordable tomato-based salsa.
- Suggested using ground turkey or chicken as a cost-effective alternative to ground beef.

**User Prompt**: Nice. How would you modify the recipe to account for seasonal summer availability of certain foods in the Northern mid-latitudes?

**Assistant Response**:
**Title:** Summer Seasonal Surf 'n' Turf Tacos with Fresh Corn & Berry Salsa
Approximate Time to Complete: 35 minutes
Serves: 2 people

**Ingredients:**
- 1/2 medium onion, finely chopped
- 1/2 cup fresh corn kernels, cut from the cob
- 1/2 cup milk
- 1/2 tablespoon sugar
- 1/2 cup shredded cheese (of your choice)
- 1/2 pound ground beef (or ground turkey or chicken for a lighter option)
- 1/2 cup flour
- 1/2 cup mushrooms, sliced
- 4 small flour tortillas

Fresh Corn & Berry Salsa:
- 1/2 cup diced strawberries
- 1/2 cup diced fresh tomatoes
- 1/4 cup finely chopped onion
- 1/2 jalapeño, seeded and finely chopped (optional)

|  |
|---|
| • 1/2 cup fresh corn kernels<br>• Salt and pepper, to taste<br>**Instructions:**<br>1. In a medium bowl, combine diced strawberries, tomatoes, 1/4 cup of the chopped onion, jalapeño (if using), fresh corn, salt, and pepper. Mix well and set aside to let the flavors meld, creating the fresh corn and berry salsa.<br>2. In a large skillet, cook the ground beef (or ground turkey or chicken) over medium heat until browned. Drain the excess fat and set aside.<br>3. In the same skillet, heat a small amount of oil over medium heat. Cook the mushrooms, remaining onion, and fresh corn kernels until the vegetables are tender. Set aside.<br>4. In a small saucepan, heat the milk over low heat. Gradually whisk in the flour to create a thick, smooth sauce. Stir in the shredded cheese until melted and combined. Remove from heat.<br>5. Warm the flour tortillas in a dry skillet or microwave for a few seconds until they are pliable.<br>6. To assemble the tacos, spread a thin layer of the cheese sauce on each tortilla. Add a portion of the ground meat, mushroom-onion-corn mixture, and drizzle with a bit more cheese sauce.<br>7. Top the tacos with the fresh corn and berry salsa and serve immediately. Enjoy your Summer Seasonal Surf 'n' Turf Tacos with Fresh Corn & Berry Salsa!<br>**Modifications to account for seasonal summer availability of certain foods in the Northern mid-latitudes:**<br>• Incorporated fresh corn kernels in both the tacos and the salsa, as corn is abundant in the summer.<br>• Replaced the original berry salsa with a fresh corn and berry salsa that includes strawberries and fresh tomatoes, which are also more readily available in the summer.<br>• Suggested using ground turkey or chicken as an alternative to ground beef for a lighter, more summer-appropriate option. |
| **User Prompt**: Nice. How would you modify the recipe to account for seasonal winter availability of certain foods in the Northern mid-latitudes? |
| **Assistant Response**:<br>**Title:** Winter Seasonal Surf 'n' Turf Tacos with Root Vegetable Salsa<br>Approximate Time to Complete: 35 minutes<br>Serves: 2 people<br>**Ingredients:**<br>• 1/2 medium onion, finely chopped<br>• 1/2 cup frozen corn<br>• 1/2 cup milk<br>• 1/2 tablespoon sugar<br>• 1/2 cup shredded cheese (of your choice)<br>• 1/2 pound ground beef (or ground turkey or chicken for a lighter option)<br>• 1/2 cup flour<br>• 1/2 cup mushrooms, sliced<br>• 4 small flour tortillas<br>Root Vegetable Salsa:<br>• 1/2 cup diced carrots, parboiled<br>• 1/2 cup diced parsnips, parboiled<br>• 1/4 cup finely chopped onion<br>• 1/2 jalapeño, seeded and finely chopped (optional)<br>• Salt and pepper, to taste<br>**Instructions:**<br>1. In a medium bowl, combine diced carrots, parsnips, 1/4 cup of the chopped onion, jalapeño (if using), salt, and pepper. Mix well and set aside to let the flavors meld, creating the root vegetable salsa.<br>2. In a large skillet, cook the ground beef (or ground turkey or chicken) over medium heat until browned. Drain the excess fat and set aside.<br>3. In the same skillet, heat a small amount of oil over medium heat. Cook the mushrooms, remaining onion, and frozen corn until the vegetables are tender. Set aside.<br>4. In a small saucepan, heat the milk over low heat. Gradually whisk in the flour to create a thick, smooth sauce. Stir in the shredded cheese until melted and combined. Remove from heat. |

5. Warm the flour tortillas in a dry skillet or microwave for a few seconds until they are pliable.
6. To assemble the tacos, spread a thin layer of the cheese sauce on each tortilla. Add a portion of the ground meat, mushroom-onion-corn mixture, and drizzle with a bit more cheese sauce.
7. Top the tacos with the root vegetable salsa and serve immediately. Enjoy your Winter Seasonal Surf 'n' Turf Tacos with Root Vegetable Salsa!

**Modifications to account for seasonal winter availability of certain foods in the Northern mid-latitudes:**
- Replaced fresh corn with frozen corn, as fresh corn is less available in winter.
- Removed the summer-inspired fresh corn and berry salsa and replaced it with a root vegetable salsa featuring parboiled carrots and parsnips, which are more readily available in winter.
- Maintained the option to use ground turkey or chicken as an alternative to ground beef for a lighter option, which can still be appropriate for winter meals.

**User Prompt**: Nice. Modify the original brunch recipe to minimize cooking and preparation time because customers are busy and have to get back to work

**Assistant Response**:
**Title:** Quick Surf 'n' Turf Tacos with Berry Salsa
Approximate Time to Complete: 20 minutes
Serves: 2 people
**Ingredients:**
- 1/2 medium onion, finely chopped
- 1/2 cup frozen corn, thawed
- 1/2 cup store-bought cheese sauce
- 1/2 pound pre-cooked shrimp, peeled and deveined
- 1/2 pound pre-cooked ground beef (or ground turkey or chicken for a lighter option)
- 1/2 cup pre-sliced mushrooms
- 4 small flour tortillas

Quick Berry Salsa:
- 1/2 cup store-bought salsa
- 1/2 cup diced strawberries
- 1/2 cup blueberries

**Instructions:**
1. In a medium bowl, combine store-bought salsa, strawberries, and blueberries. Mix well and set aside to let the flavors meld, creating the quick berry salsa.
2. In a large skillet, heat a small amount of oil over medium heat. Cook the pre-sliced mushrooms, chopped onion, and thawed corn until the vegetables are tender. Set aside.
3. Warm the flour tortillas in a dry skillet or microwave for a few seconds until they are pliable.
4. To assemble the tacos, spread a thin layer of store-bought cheese sauce on each tortilla. Add a portion of the pre-cooked ground meat, shrimp, and mushroom-onion-corn mixture. Drizzle with a bit more cheese sauce.
5. Top the tacos with the quick berry salsa and serve immediately. Enjoy your Quick Surf 'n' Turf Tacos with Berry Salsa!

**Modifications to minimize cooking and preparation time:**
- Replaced fresh corn with thawed frozen corn to save time on preparation.
- Used store-bought cheese sauce instead of making it from scratch.
- Utilized pre-cooked shrimp and ground beef (or ground turkey or chicken) to eliminate cooking time.
- Chose pre-sliced mushrooms to save time on slicing.
- Modified the berry salsa by combining store-bought salsa with strawberries and blueberries for a faster preparation.

**User Prompt**: Nice. Modify the previous recipe for vegans

**Assistant Response**:
**Title:** Quick Vegan Veggie Tacos with Berry Salsa
Approximate Time to Complete: 20 minutes
Serves: 2 people
**Ingredients:**
- 1/2 medium onion, finely chopped

- 1/2 cup frozen corn, thawed
- 1/2 cup store-bought vegan cheese sauce (or cashew cream sauce)
- 1 cup canned black beans, drained and rinsed
- 1/2 cup pre-sliced mushrooms
- 4 small flour tortillas (make sure they are vegan)

Quick Berry Salsa:
- 1/2 cup store-bought salsa
- 1/2 cup diced strawberries
- 1/2 cup blueberries

**Instructions:**
1. In a medium bowl, combine store-bought salsa, strawberries, and blueberries. Mix well and set aside to let the flavors meld, creating the quick berry salsa.
2. In a large skillet, heat a small amount of oil over medium heat. Cook the pre-sliced mushrooms, chopped onion, and thawed corn until the vegetables are tender. Set aside.
3. Warm the flour tortillas in a dry skillet or microwave for a few seconds until they are pliable.
4. To assemble the tacos, spread a thin layer of store-bought vegan cheese sauce (or cashew cream sauce) on each tortilla. Add a portion of the canned black beans and mushroom-onion-corn mixture. Drizzle with a bit more vegan cheese sauce (or cashew cream sauce).
5. Top the tacos with the quick berry salsa and serve immediately. Enjoy your Quick Vegan Veggie Tacos with Berry Salsa!

**Modifications to make the recipe vegan:**
- Replaced shrimp and ground meat with canned black beans as a vegan protein source.
- Used store-bought vegan cheese sauce or cashew cream sauce instead of traditional cheese sauce.
- Ensured that the flour tortillas used are vegan (without any animal-derived ingredients such as lard).

**User Prompt**: Nice. Modify the original recipe for keto. Remember the ingredients list had ingredient list of onion, strawberries, corn, milk, sugar, cheese, banana, shrimp, goat_cheese, blueberries, ground_beef, flour, mushrooms

**Assistant Response**:
**Title:** Keto-Friendly Surf 'n' Turf Tacos with Goat Cheese and Berry Salsa
Approximate Time to Complete: 30 minutes
Serves: 2 people
**Ingredients:**
- 1/2 medium onion, finely chopped
- 1/2 cup shrimp, peeled and deveined
- 1/2 pound ground beef (or ground turkey or chicken for a lighter option)
- 1/2 cup crumbled goat cheese
- 1/2 cup sliced mushrooms
- 4 large lettuce leaves (such as iceberg or butter lettuce) or keto tortillas

Berry Salsa (low-carb version):
- 1/4 cup diced strawberries
- 1/4 cup blueberries
- 1/4 cup finely chopped onion
- 1/2 jalapeño, seeded and finely chopped (optional)
- Salt and pepper, to taste

**Instructions:**
1. In a medium bowl, combine diced strawberries, blueberries, 1/4 cup of the chopped onion, jalapeño (if using), salt, and pepper. Mix well and set aside to let the flavors meld, creating the low-carb berry salsa.
2. In a large skillet, cook the ground beef (or ground turkey or chicken) over medium heat until browned. Drain the excess fat and set aside.
3. In the same skillet, cook the shrimp until pink and cooked through. Remove from the skillet and set aside.
4. In the same skillet, heat a small amount of oil over medium heat. Cook the mushrooms and remaining onion until the vegetables are tender. Set aside.
5. If using keto tortillas, warm them in a dry skillet or microwave for a few seconds until they are pliable. If using lettuce leaves, rinse and pat them dry.

6. To assemble the tacos, place a layer of crumbled goat cheese on each lettuce leaf or keto tortilla. Add a portion of the ground meat, shrimp, and mushroom-onion mixture.
7. Top the tacos with the low-carb berry salsa and serve immediately. Enjoy your Keto-Friendly Surf 'n' Turf Tacos with Goat Cheese and Berry Salsa!

**Modifications to make the original recipe keto-friendly:**
- Removed corn, milk, sugar, flour, and banana from the recipe, as they are high in carbs and not keto-friendly.
- Replaced regular tortillas with keto tortillas or lettuce leaves as a low-carb alternative.
- Adjusted the berry salsa recipe to use only strawberries and blueberries, which are lower in carbs than most fruits.
- Used goat cheese instead of a cheese sauce to minimize the use of milk and flour.

**User Prompt**: Nice. Modify the original recipe for serving 3 patrons who are lactose-intolerant

**Assistant Response**:
**Title:** Lactose-Free Surf 'n' Turf Tacos with Fruit Salsa
Approximate Time to Complete: 35 minutes
**Serves: 3 people**
**Ingredients:**
- 1 medium onion, finely chopped
- 1/2 cup diced strawberries
- 1/2 cup blueberries
- 1/2 pound shrimp, peeled and deveined
- 1/2 pound ground beef (or ground turkey or chicken for a lighter option)
- 1/2 cup sliced mushrooms
- 6 small flour tortillas

Fruit Salsa:
- 1/2 cup diced strawberries
- 1/2 cup blueberries
- 1 banana, diced
- 1/4 cup finely chopped onion
- 1/2 jalapeño, seeded and finely chopped (optional)
- Salt and pepper, to taste

**Instructions:**
1. In a medium bowl, combine diced strawberries, blueberries, banana, 1/4 cup of the chopped onion, jalapeño (if using), salt, and pepper. Mix well and set aside to let the flavors meld, creating the fruit salsa.
2. In a large skillet, cook the ground beef (or ground turkey or chicken) over medium heat until browned. Drain the excess fat and set aside.
3. In the same skillet, cook the shrimp until pink and cooked through. Remove from the skillet and set aside.
4. In the same skillet, heat a small amount of oil over medium heat. Cook the mushrooms and remaining onion until the vegetables are tender. Set aside.
5. Warm the flour tortillas in a dry skillet or microwave for a few seconds until they are pliable.
6. To assemble the tacos, add a portion of the ground meat, shrimp, and mushroom-onion mixture to each tortilla.
7. Top the tacos with the fruit salsa and serve immediately. Enjoy your Lactose-Free Surf 'n' Turf Tacos with Fruit Salsa!

**Modifications for lactose-intolerant patrons:**
- Removed cheese, goat cheese, milk, and cheese sauce from the recipe to make it lactose-free.
- Introduced a fruit salsa that includes banana, strawberries, and blueberries for added flavor in the absence of dairy ingredients.
- Adjusted the number of tortillas and increased the ingredient amounts to accommodate serving 3 people.

**User Prompt**: Nice. Modify the original recipe for a home cook to provide dinner for two and a follow-up breakfast starter from leftovers for 1 person

**Assistant Response**:
**Title:** Surf 'n' Turf Tacos with Berry Salsa Dinner for Two and Breakfast Scramble for One
Approximate Time to Complete: 35 minutes (Dinner) + 10 minutes (Breakfast)

**Dinner Serves: 2 people Breakfast Serves: 1 person**
**Ingredients:**
- 1 medium onion, finely chopped
- 1/2 cup diced strawberries
- 1/2 cup blueberries
- 1/2 pound shrimp, peeled and deveined
- 1/2 pound ground beef (or ground turkey or chicken for a lighter option)
- 1/2 cup shredded cheese
- 1/2 cup sliced mushrooms
- 1/2 cup milk
- 1 tablespoon sugar
- 4 small flour tortillas (Dinner)
- 2 large eggs (Breakfast)
- Salt and pepper, to taste

Berry Salsa:
- 1/2 cup diced strawberries
- 1/2 cup blueberries
- 1/4 cup finely chopped onion
- 1/2 jalapeño, seeded and finely chopped (optional)
- Salt and pepper, to taste

**Instructions for Dinner:**
1. In a medium bowl, combine diced strawberries, blueberries, 1/4 cup of the chopped onion, jalapeño (if using), salt, and pepper. Mix well and set aside to let the flavors meld, creating the berry salsa.
2. In a large skillet, cook the ground beef (or ground turkey or chicken) over medium heat until browned. Drain the excess fat and set aside. Reserve half of the cooked ground meat for breakfast.
3. In the same skillet, cook the shrimp until pink and cooked through. Remove from the skillet and set aside.
4. In the same skillet, heat a small amount of oil over medium heat. Cook the mushrooms and remaining onion until the vegetables are tender. Set aside.
5. Warm the flour tortillas in a dry skillet or microwave for a few seconds until they are pliable.
6. To assemble the tacos, add a portion of the ground meat, shrimp, and mushroom-onion mixture to each tortilla. Sprinkle with shredded cheese.
7. Top the tacos with the berry salsa and serve immediately.

**Instructions for Breakfast:**
1. In a small bowl, whisk together the two large eggs, milk, sugar, salt, and pepper.
2. In a non-stick skillet, heat a small amount of oil over medium heat. Add the reserved cooked ground meat from dinner and cook for 1-2 minutes until heated through.
3. Pour the egg mixture over the ground meat in the skillet. Cook, stirring occasionally, until the eggs are fully cooked and scrambled.
4. Serve the breakfast scramble with a side of the leftover berry salsa. Enjoy your Surf 'n' Turf Tacos with Berry Salsa Dinner and Breakfast Scramble!

**Modifications to provide dinner for two and a follow-up breakfast starter for one:**
- Reserved half of the cooked ground meat for the breakfast scramble.
- Created a breakfast scramble recipe that uses the reserved ground meat, eggs, milk, sugar, salt, and pepper.
- Served the breakfast scramble with leftover berry salsa from the dinner recipe.

**User Prompt**: Nice. Modify the original recipe for one person and one take-home doggy bag intended for a poodle. as leftovers. Take note of ingredients that may be inappropriate for small canines.

**Assistant Response:**
**Title:** Surf 'n' Turf Tacos with Berry Salsa for One and a Doggy Bag for a Poodle
Approximate Time to Complete: 35 minutes
**Serves: 1 person + 1 poodle**
**Ingredients:**
- 1/2 medium onion, finely chopped
- 1/4 cup diced strawberries
- 1/4 cup blueberries

- 1/4 pound shrimp, peeled and deveined
- 1/4 pound ground beef (or ground turkey or chicken for a lighter option)
- 1/4 cup shredded cheese
- 1/4 cup sliced mushrooms
- 2 small flour tortillas (for the person)

Berry Salsa:
- 1/4 cup diced strawberries
- 1/4 cup blueberries
- 2 tablespoons finely chopped onion
- Salt and pepper, to taste

**Instructions for Dinner for One:**
1. In a medium bowl, combine diced strawberries, blueberries, 2 tablespoons of the chopped onion, salt, and pepper. Mix well and set aside to let the flavors meld, creating the berry salsa.
2. In a large skillet, cook the ground beef (or ground turkey or chicken) over medium heat until browned. Drain the excess fat and set aside. Reserve half of the cooked ground meat for the doggy bag.
3. In the same skillet, cook the shrimp until pink and cooked through. Remove from the skillet and set aside. Reserve one shrimp for the doggy bag.
4. In the same skillet, heat a small amount of oil over medium heat. Cook the mushrooms and remaining onion until the vegetables are tender. Set aside.
5. Warm the flour tortillas in a dry skillet or microwave for a few seconds until they are pliable.
6. To assemble the taco for one person, add a portion of the ground meat, shrimp, and mushroom-onion mixture to the tortillas. Sprinkle with shredded cheese.
7. Top the taco with the berry salsa and serve immediately.

**Instructions for the Doggy Bag for a Poodle:**
1. In a separate container, combine the reserved cooked ground meat and one cooked shrimp. Do not add any onions, as they can be toxic to dogs.
2. Add a small portion of cooked mushrooms, avoiding any onions that might be mixed in.
3. Do not add any cheese or berry salsa to the doggy bag, as some dogs may have trouble digesting dairy products, and the salsa contains onions.
4. When serving the doggy bag to your poodle, make sure the leftovers are at room temperature or slightly warmed, not hot.

**Modifications for one person and a doggy bag for a poodle:**
- Divided the ingredients in half to create a smaller serving size for one person.
- Reserved half of the cooked ground meat and one cooked shrimp for the doggy bag.
- Removed onions, cheese, and berry salsa from the doggy bag to accommodate a poodle's dietary needs.
- Provided instructions for serving the doggy bag leftovers to a poodle.